# A Fuzzy Relational Identification Algorithm and Its Application to Predict the Behaviour of a Motor Drive System

P.J. Costa Branco and J.A. Dente

*Instituto Superior Técnico (I.S.T.)*
*Laboratório de Mecatrónica*
*Av. Rovisco Pais, 1096, Lisboa Codex*
*Portugal*
*E-mail: pbranco@alfa.ist.utl.pt*

**Abstract**

*Fuzzy relational identification builds a relational model describing system's behaviour by a nonlinear mapping between its variables. In this paper, we propose a new fuzzy relational algorithm based on simplified max-min relational equation. The algorithm presents an adaptation method applied to gravity-center of each fuzzy set based on error integral value between measured and predicted system's output, and uses the concept of time-variant universe of discourses. The identification algorithm also includes a method to attenuate noise influence in extracted system's relational model using a fuzzy filtering mechanism. The algorithm is applied to one-step forward prediction of a simulated and experimental motor drive system. The identified model has its input-output variables (stator-reference current and motor speed signal) treated as fuzzy sets, whereas the relations existing between them are described by means of a matrix R defining the relational model extracted by the algorithm. The results show the good potentialities of the algorithm in predict the behaviour of the system and attenuate through the fuzzy filtering method possible noise distortions in the relational model.*

# I. INTRODUCTION

Fuzzy systems are usually named *model-free estimators*. They estimate input-output relations without the need of an analytical model of how outputs depend on inputs, and encode the sampled information in a parallel-distributed framework called *fuzzy structure*. Three main types of fuzzy structures have been presented in the literature:

- Rule-based systems [8]: this model type is used in fuzzy control where the process behaviour is described by a set of IF-THEN rules;

- Fuzzy relational systems [6,10,11,12,16]: these systems represent an alternative to rule-based systems preserving their qualitative characteristics (can be interpreted by IF-THEN statements too) but avoiding the need for *hard-work rules* development;

- Fuzzy functional systems [14]: these systems have a hybrid structure with a linear function expressing the input-output state relationships in conclusion part, and fuzzy sets representing data in condition part.

This paper uses fuzzy relational models. These can represent nonlinear dynamical systems and can perform nonlinear mappings [2,3,13,15]. In relational models, input-output variables are related by a fuzzy relationship through a relational matrix $\boldsymbol{R}$. This matrix contains every possible combination of input-output conditions with one value for each condition between zero and one, and it can be interpreted by a rule-based approach. Each value represents the degree of truth of each possible relationship. A value of one indicates that the relationship is the strongest, and a value of zero indicates that the relationship is the weakest. The identification problem consists then in estimate the unknown relational matrix $\boldsymbol{R}$ to build a fuzzy model of the considered system.

At this point, some problems related with the identification of relational models, as well their application, can be mentioned:

- the dimension of the relational matrix ***R***. For a system having more than one input, the computational and memory effort involved in finding the relation may be too high for practical applications;
- the I/O data used for composing the fuzzy relation need to be sufficient "rich" to excite the process. Then, on-line identification becomes impracticable;
- at last, it is necessary to know possible delays in the input variables which is a hard task to more complex processes.

Taking into acount the previous drawbacks, this paper investigates a new fuzzy relational identification algorithm which initial version was presented in [1] where it was compared with other used ones. We analyse also in this paper its application in predicting a motor drive system's behaviour although it is a general algorithm to on-line modelling other processes.

The algorithm includes a technique to attenuate noise influence in extracted relational model using a fuzzy filtering mechanism. Its steps are explained and its use in the motor drive system is demonstrated with simulated and experimental results.

In Section II, we briefly review the basic fuzzy-logic operations used in this paper to develop the relational identification algorithm. Section III defines a fuzzy relational system. The noise attenuation process is described in Section IV. Section V describes the identification algorithm. At last, Section VI shows the application of the fuzzy relational algorithm in modelling a motor drive system.

## II. OVERVIEW OF THE BASIC FUZZY SET OPERATIONS

As for traditional set theory, we can define basic fuzzy set operations. Some of these operations are described below, mainly those used by the proposed identification algorithm. To illustrate each operation we use a simple numerical example.

Let $\overline{A}$ and $\overline{B}$ be two fuzzy sets defined in $X$ with membership functions $\mu_{\overline{A}}(x)$ and $\mu_{\overline{B}}(x)$, respectively.

*Definition 1: Union:* The membership function $\mu_{\overline{A}\cup\overline{B}}(x)$ of union $\overline{A}\cup\overline{B}$ is defined for all $x \in X$ by

$$\mu_{\overline{A}\cup\overline{B}}(x) = \max\{\mu_{\overline{A}}(x), \mu_{\overline{B}}(x)\}. \tag{1}$$

*Definition 2: Intersection:* The membership function $\mu_{\overline{A}\cap\overline{B}}(x)$ of intersection $\overline{A}\cap\overline{B}$ is defined for all $x \in X$ by

$$\mu_{\overline{A}\cap\overline{B}}(x) = \min\{\mu_{\overline{A}}(x), \mu_{\overline{B}}(x)\}. \tag{2}$$

*Definition 3: Complement:* The membership function $\overline{\mu}_{\overline{A}}(x)$ of the complement of a fuzzy set $\overline{A}$ is defined for all $x \in X$ by

$$\overline{\mu}_{\overline{A}}(x) = 1 - \mu_{\overline{A}}(x). \tag{3}$$

*Definition 4: Cartesian Product:* If $\overline{A}_1, \cdots, \overline{A}_n$, are fuzzy sets in $X_1, \cdots, X_n$, respectively, the Cartesian product of $\overline{A}_1, \cdots, \overline{A}_n$, is a fuzzy set in the product space $X_1 \times \cdots \times X_n$ with its membership function defined by (4) if the *min* operator is used,

$$\mu_{\overline{A}_1 \times \cdots \times \overline{A}_n}(x_1, x_2, \cdots, x_n) = \min\left\{\mu_{\overline{A}_1}(x_1), \cdots, \mu_{\overline{A}_n}(x_n)\right\}, \tag{4}$$

or, if the *product* operator is used, the membership function is defined as (5).

$$\mu_{\overline{A}_1 \times \cdots \times \overline{A}_n}(x_1, x_2, \cdots, x_n) = \mu_{\overline{A}_1}(x_1) \cdot \mu_{\overline{A}_2}(x_2) \cdots \cdot \mu_{\overline{A}_n}(x_n) \tag{5}$$

*A. Example 1*

$$\mu_{\overline{A}_1}(x) = [0.3\ 0.9\ 0.1], \quad \mu_{\overline{A}_2}(x) = [0.9\ 0.5\ 0.1] \tag{6}$$

$$\mu_{\overline{A}_1} \times \mu_{\overline{A}_2} = \min\left\{\mu_{\overline{A}_1}(x), \mu_{\overline{A}_2}(x)\right\} \tag{7}$$

$$\mu_{\overline{A}_1} \times \mu_{\overline{A}_2} = \begin{bmatrix} 0.3 & 0.3 & 0.1 \\ 0.9 & 0.5 & 0.1 \\ 0.1 & 0.1 & 0.1 \end{bmatrix} \tag{8}$$

*Definition 5: Fuzzy Relation:* An $n$-ary fuzzy relation is a fuzzy set in $X_1 \times \cdots \times X_n$ expressed as

$$\boldsymbol{R}_{X_1 \times \cdots \times X_n} = \left\{\left((x_1, \cdots, x_n), \mu_R(x_1, \cdots, x_n)\right) \middle| (x_1, \cdots, x_n) \in X_1 \times \cdots \times X_n\right\} \tag{9}$$

*B. Example 2*

In example 1, the Cartesian product $\mu_{\overline{A}_1} \times \mu_{\overline{A}_2}$ composes a relation between fuzzy sets $\overline{A}_1$ and $\overline{A}_2$ in the following form:

$$\boldsymbol{R}_{\mu_{\overline{A}_1} \times \mu_{\overline{A}_2}} \Rightarrow \begin{matrix} & x_{21} & x_{22} & x_{23} \\ x_{11} & 0.3 & 0.3 & 0.1 \\ x_{12} & 0.9 & 0.5 & 0.1 \\ x_{13} & 0.1 & 0.1 & 0.1 \end{matrix}. \tag{10}$$

An element of this relation $\boldsymbol{R}$ can be interpreted as, for example,

$$[\text{IF } (x_{11} \textit{ is big}) \text{ THEN } (x_{21} \textit{ is small})] \textit{ with possibility of } 0.3.$$

*Definition 6: Sup-Star Compositional Rule of Inference:* If $\boldsymbol{R}$ is a fuzzy relation in $X \times Y$, and $\overline{X}$ is a fuzzy set in *X*, then the "sup-star compositional rule of inference" defines that the fuzzy set $\overline{Y}$ in *Y* induced by $\overline{X}$ in $\boldsymbol{R}$ is given by

$$\overline{Y} = \overline{X} \bullet \boldsymbol{R}. \tag{11}$$

The symbol "•" represents the *sup-star composition* and it can be any operator in the class of triangular norms, for example, minimum, algebraic product, bounded product, or drastic product. If the minimum operator is used, then this definition reduces to Zadeh's compositional rule of inference.

*C. Example 3*

Let

$$\overline{X} = \begin{bmatrix} 0.2 & 1.0 & 0.3 \end{bmatrix}, \tag{12}$$

$$\boldsymbol{R} = \begin{bmatrix} 0.8 & 0.9 & 0.2 \\ 0.6 & 1.0 & 0.4 \\ 0.5 & 0.8 & 1.0 \end{bmatrix}. \tag{13}$$

Therefore, the fuzzy set $\overline{Y}$ is induced by $\overline{X}$ in $\boldsymbol{R}$ as

$$\overline{Y} = \begin{bmatrix} 0.2 & 1.0 & 0.3 \end{bmatrix} \bullet \begin{bmatrix} 0.8 & 0.9 & 0.2 \\ 0.6 & 1.0 & 0.4 \\ 0.5 & 0.8 & 1.0 \end{bmatrix} = \begin{bmatrix} 0.6 & 1.0 & 0.4 \end{bmatrix} \tag{14}$$

## III. FUZZY RELATIONAL SYSTEMS

The systems under study are modeled as the fuzzy systems described in (15). In this equation, the symbol "•" represents the *max-min* composition operator, $\overline{X}_{1,} \overline{X}_{2,} \ldots, \overline{X}_n$ denote the input fuzzy sets, $\overline{Y}$ stands for the output fuzzy set, and $\boldsymbol{R}$ is the fuzzy relational matrix expressing the system's input-output relationship.

$$\overline{Y} = \overline{X}_1 \bullet \overline{X}_2 \bullet \ldots \bullet \overline{X}_n \bullet \boldsymbol{R} \tag{15}$$

Fuzzy relational equation (15) describes multiple-input single-output fuzzy systems. From a system theory point of view, the following simplified version of (15) can be considered as a single-input single-output fuzzy system

$$\overline{Y} = \overline{X} \bullet \boldsymbol{R}, \tag{16}$$

and (17) its discretized version for each instant *k*.

$$\overline{Y}_k = \overline{X}_k \bullet \boldsymbol{R}_k \tag{17}$$

Equation (17) can also be rewritten as

$$\overline{Y}_k(y_k) = \sup_{x_k \in X} \left[ \min\left( \overline{X}_k(x_k), \boldsymbol{R}_k(x_k, y_k) \right) \right]. \tag{18}$$

A fuzzy relation $\boldsymbol{R}$ is written as a set of fuzzy rules with fuzzy sets defined on each universe of discourse. For a single-input single-output system (16) defined with *n* fuzzy sets for $\overline{X}$ and $\overline{Y}$, $\boldsymbol{R}$ is an $n \times n$ matrix of possibility measures being each element denoted as in expression (19).

$$\boldsymbol{R}(i, j) = p_{ij}. \tag{19}$$

Each matrix element can be translated as a linguistic simple rule like

$$\left[ \text{IF } \overline{X}_i \text{ THEN } \overline{Y}_j \right] \textit{ with possibility } p_{ij}, \tag{20}$$

and for each condition $\overline{X}_i$ there are *n* simple rules that form a called compound rule.

This section described the main fuzzy set operations employed by the proposed identification algorithm, and defined the structure of the fuzzy relational system. In the next section, we explain the use of a fuzzy filtering mechanism to attenuate the effects of noise on extracted relational model.

## IV. NOISE ATTENUATION BY FUZZY FILTERING

The noise attenuation is processed based on conventional smoothing and filtering problem extended to a fuzzy situation. The objective of this *fuzzy filter* is attenuate the noise influence from estimated relations of matrix ***R*** but without distorting the model. The mechanism used by the filtering process is the *exponential smoothing method* based on the geometrical modified law [7].

### *A. The Exponential Smoothing Method*

This method deals with the effect of the numerical value on a time sequence where the signal is given at time $t$ by a convolution summation, as indicated in (21), to a hypothetical signal $y$.

$$\hat{y}(t) = (1-\gamma)^{t+1} y(0) + \sum_{i=0}^{t} \gamma(1-\gamma)^{t-i} y(i) \tag{21}$$

The convolution summation process is, in a way, a *discounting* of past data, which geometrically decreases the values of the time sequence from 0 to $t$ with the accumulation of the remaining weights starting at $t$ = 0. The parameter $\gamma$ in expression (21) regulates the filter rate.

To obtain a recursive formula of (21) , we compute $(1-\gamma)y(t-1)$ by

$$\hat{y}(t)\Big|_{n=1} = (1-\gamma)y(t-1) + \gamma y(t) \tag{22}$$

that is a first order smoothing ($n = 1$). Smoothing of order $n$ are calculated by superimposing $n$ smoothings of order 1. Thus, for example, the expression (23) is a smoothing of order $n = 2$.

$$\hat{y}(t)\Big|_{n=2} = (1-\gamma)\hat{y}(t-1)\Big|_{n=1} + \gamma\hat{y}(t)\Big|_{n=1} \tag{23}$$

## V. THE RELATIONAL IDENTIFICATION ALGORITHM

The identification algorithm is divided in two stages :

- first stage predicts the system output using the initial relational model, and adjusts the gravity-center of each fuzzy set using the integral of error value between predicted and measured output values;
- second stage actualises the relational matrix using new input-output system's data values, and applies the previous explained fuzzy filtering process to each model relation attenuating noise influence in the extracted relational model.

Completing the algorithm, we introduce the concept of *variable universes of discourse*.

### *A. The Gravity-Center Adjustment Method*

Membership functions characterising the input-output system variables are presented in figure 2 by 7 fuzzy sets and their respective gravity-centers (CG). On the basis of these fuzzy sets, numerical values are translated from linguistic (24) to numerical representation (25) by the discretized gravity-center method. The inferred numerical value $\hat{y}$ results then in a media of linguistic values weighted by the gravity-center of each referential set.

$$\overline{Y} = [NB\ NM \cdots\ PB] \tag{24}$$

$$\hat{y} = \frac{NB.CG_{NB} + NM.CG_{NM} + \cdots + PB.CG_{PB}}{CG_{NB} + CG_{NM} + \cdots + CG_{PB}} \tag{25}$$

The gravity-center adjustment method is based on integral of the error signal $e(k)$ in (26) between measured systems output value $y(k)$ and its predicted value $\hat{y}(k)$ by the fuzzy relational model. The adjustment of each $i$ gravity-center $(CG_i)$, as indicated in expression (27), is made in the same direction of the integral error value, and it is proportional to the respective estimated linguistic value $\hat{\bar{Y}}_i(k)$ of the last estimated fuzzy output vector $\hat{\bar{Y}}(k)$. Using this procedure, the output signal dynamics can be incorporated into the fuzzy sets partition in a way to compensate the use of a bidimensional matrix that does not have information from past output signals, thus helping to reduce the error between predicted and real signals.

$$e(k) = y(k) - \hat{y}(k). \tag{26}$$

$$CG_i(k+1) = CG_i(k) + \left(\alpha \cdot \int e(k) \cdot \hat{\bar{Y}}_i(k)\right) \tag{27}$$

$$i = NB, \ldots, PB$$

In (27), the parameter $\alpha$ is responsible for the adjusting rate of the gravity-centers. To values of $\alpha >> 1$, the identification algorithm can present oscillations in the predicted values because gravity-center's adjustment becomes faster than system output dynamics. Otherwise, for very small $\alpha$ values, the algorithm’s performance deteriorates with the error between measured and predicted values growing up.

*B. The Concept of Time-Variant Universes of Discourse*

Some proposed identification algorithms in literature present a "saturation effect" in predicted values near from limits of the universe of discourse. This happens because the algorithm has insufficient information when working in the universe of discourse limits coming from non-optimal membership functions. To overcome this effect, we propose a variable percentage expansion β in the initial limit values of the universes of discourse. Therefore, as the input-output system values come near from universe of discourse limits, these are expanded until the new limit values, determined by parameter β, are reached.

Suppose that variables *UInic* and *YInic* are, respectively, the initial values of the input and output universes of discourse limits. Define $UnivDisc1(k)$ and $UnivDisc2(k)$ as the boundary values for input-output universes of discourse limits in time instant *k*. The expansion is processed at same time in the two universes and it is described in equations (28) and (29).

$$UnivDisc1(k) = UInic.\left[1.0 + \left(\frac{\beta.|u(k)|}{UInic}\right)\right] \tag{28}$$

$$UnivDisc2(k) = UInic.\left[1.0 + \left(\frac{\beta.|y(k)|}{UInic}\right)\right] \tag{29}$$

*C. The Identification Algorithm*

Following, we explain the proposed relational identification algorithm based on the steps shown in the block diagram at figure 3.

Suppose three time instants in the identification process, $(k-1)$, $(k)$, and $(k+1)$. Each data pair $(x(k), y(k))$ represents the system input-output values having their correspondent fuzzy values, $\overline{X}_k$ and $\overline{Y}_k$, obtained from the adjusted fuzzy sets in iteration $(k-1)$ using the gravity-center adjustment equation (27). The relational matrix at instant *k*th, $\boldsymbol{R}_k$, is obtained by (30). This expression computes the fuzzy union between Cartesian product of new fuzzy vectors $\overline{X}_k$ and $\overline{Y}_k$, with the relational matrix $\boldsymbol{R}_{(k-1)}$ obtained from previous time step $(k-1)$.

$$\boldsymbol{R}_k = \left(\overline{X}_k \times \overline{Y}_k\right) \cup \boldsymbol{R}_{(k-1)} \tag{30}$$

Following, equation (31) predicts one-step forward the fuzzy system output $\hat{\bar{Y}}_{k+1}$ using the next input signal $\bar{X}_{k+1}$ and its *max-min* composition with the relational matrix $\boldsymbol{R}_k$ computed in (30).

$$\hat{\bar{Y}}_{k+1} = \bar{X}_{k+1} \bullet \boldsymbol{R}_k . \tag{31}$$

The estimated fuzzy output value $\hat{\bar{Y}}_{k+1}$ is "translated" to its numerical value $\hat{y}_{k+1}$ by the gravity-center method. The numerical value is compared with the real one, $y_{k+1}$, to generate the error signal (32) used by the gravity-center adjustment mechanism to "calibrate" the membership functions using relation (27).

$$e_{k+1} = y_{k+1} - \hat{y}_{k+1} \tag{32}$$

The second stage of the algorithm begins after the gravity-center adjustment process be computed. Using measured fuzzy states $\bar{X}_{k+1}$ and $\bar{Y}_{k+1}$, a "measured" relational matrix $\boldsymbol{R}'_{(k+1)}$ is constructed as indicated in (33).

$$\boldsymbol{R}'_{k+1} = \bar{X}_{k+1} \times \bar{Y}_{k+1} \tag{33}$$

The final relational matrix for $(k+1)$ instant, $\boldsymbol{R}_{k+1}$, arises from the fuzzy union (34) between previous relation $\boldsymbol{R}_k$ in (30), and the measured relation $\boldsymbol{R}'_{k+1}$ expressed in equation (33). The final relational matrix $\boldsymbol{R}_{k+1}$ in (35) is then composed by the strongest input-output relations between "measured" and "corrected" relation matrixes.

$$\boldsymbol{R}_{k+1} = \boldsymbol{R}'_{k+1} \cup \boldsymbol{R}_k \tag{34}$$

$$\boldsymbol{R}_{k+1} = \left( \bar{X}_{k+1} \times \bar{Y}_{k+1} \right) \cup \boldsymbol{R}_k . \tag{35}$$

The final matrix $\boldsymbol{R}_{k+1}$ is now *filtered* using the exponential smoothing method in each matrix element attenuating possible noise contamination of extracted relational model.

The iteration cycle of the indentification algorithm ends with the *filtered* matrix $\boldsymbol{R}_{k+1}$ being used in $(k+2)$ system's output prediction and so begin the algorithm for the next 3 steps $(k)$, $(k+1)$, and $(k+2)$.

## VI. RELATIONAL IDENTIFICATION OF A MOTOR DRIVE SYSTEM

In this section, we apply the identification algorithm in predicting the behaviour of a motor drive system through simulated and experimental results. Different tests are used to investigate the characteristics of the algorithm in extract the relational model.

The motor drive system is based on a permanent-magnet motor and a current-controlled inverter. The motor is represented by equation set (36) in *d*- and *q* components of the stator current and angular speed $\omega$. In Appendix, we describe each equation symbol and it is given their values.

Figure 4 shows a diagram of the system used in our simulated and experimental tests. The diagram shows a permanent-magnet motor (P.M. Motor) driven by the current-controlled inverter. Sliding-mode control principle is applied to current control, resulting in a simple and improved hysteresis current-controlled inverter that needs only two phase current measures, $i_1$ and $i_2$. For more details, references [4,5,9] describe this technique.

The motor drive system has two inputs, $iq_{ref}$ and $id_{ref}$. The inductances are considered equal, $L_d \approx L_q$. Current $id_{ref}$ is nulled to result in a decoupled system. Therefore, all system can be considered to be SISO (Single-Input Single-Output) with the input system signal being $iq_{ref}$ and the system output being the motor speed value $\omega$.

$$\begin{cases} \dfrac{di_d}{dt} = \dfrac{1}{L_d}\left(u_d + \omega \cdot L_q \cdot i_q - R \cdot i_d\right) \\ \dfrac{di_q}{dt} = \dfrac{1}{L_q}\left(u_q + \omega \cdot L_d \cdot i_d - R \cdot i_q - \omega \cdot \phi_f\right) \\ \dfrac{d\omega}{dt} = \dfrac{1}{J}\left\{\left[\left(L_d - L_q\right) \cdot i_d + \phi_f\right] \cdot i_q - T_x\right\} \\ \dfrac{d\theta}{dt} = \omega \end{cases} \tag{36}$$

First simulated results apply a low frequency sinusoidal current $iq_{ref}$ with noise superimposed, as shown in figure 5, with its magnitude covering all possible current domain from -10A to +10A. As the input signal $iq_{ref}$ is applied, the motor speed signal $\omega$ and $iq_{ref}$ itself are employed by the relational identification algorithm to extract the system's relational matrix ***R***. Values of the three algorithm parameters $\alpha$, $\beta$, and $\gamma$, used, respectively, to adjust the fuzzy sets, expand the input-output universes of discourse, and filter the noise, had in this test the values of $\alpha$ = 2.3, $\beta$ = 0.82, and $\gamma$ =0.01. The filter used a first order smoothing method $(n = 1)$. Figure 6 shows the motor speed and the predicted signal inferred from the on-line extracted relational model, while figure 7 displays the error evolution. The results show that as the relational model is being extracted, the error signal decreases because there is a better prediction of the speed signal in next step since it is based on a more complete relational model.

The previous simulated results considered only noise perturbing the input signal $iq_{ref}$. Since the motor drive system has a high constant time, the input noise is filtered and it practically does not appear on speed signal. Thus, a low filter rate as $\gamma = 0.01$ demonstrated to be enough to do a good noise attenuation.

In a second test, we consider the presence of noise on input $iq_{ref}$ and output speed signal. In this case, noise in the speed signal simulates the always present noise coming from the experimental measurements. Both noise signals were simulated by a random variable with its amplitude ranging from ±10% of nominal value for each variable which is acceptable in practice. Results in figure 8 compare the speed signal, with noise superimposed and represented by a solid line, and the predicted speed, in dotted line, after apply a *fuzzy filtering* of order $n = 1$ into the extracted relational matrix.

The parameters used by the algorithm in this test were: $\alpha = 1.4$, $\beta = 0.9$, and $\gamma = 0.15$. Comparing this parameter set with that used in the previous test, we have:

- the $\alpha$ value of 1.4 becomes smaller than the anterior value of 2.3 as we do not want that noise dynamics be present during the adjustment process of the membership functions, so introducing perturbations in the defuzzification process;
- the $\beta$ value of 0.9 has a higher value because the signals, when near the universe of discourse limits, can pass them caused by noise superposition;
- at last, as we have in this test a noise presence in the speed signal too, it is need a higher filter rate of $\gamma = 0.15$ to have a good attenuation.

Results in figure 9 use the same input-output signals as before but now applying a filter of order $n = 2$. In this case, the predicted speed signal becomes more *clean*, but the 2º order filter begins to distort the relational matrix elements deforming and retarding the predicted speed signals as shown in the figure. The algorithm parameters in this test were:

- $\alpha = 0.35$ - A value smaller than that used when it was employed a first order filter. The considerable distortion caused by the use of a second order filter provokes higher errors. Thus, with their higher integral values, it became necessary to reduce the $\alpha$ value in the manner not generate oscillations in the predicted output signal;
- $\beta = 0.9$ - The same value as before.
- $\gamma = 0.35$ - This test used a second order filter $n = 2$. The value attributed to its $\gamma$ parameter was a compromise to obtain a minimum distortion without great noise influence in the extracted relational model;

In the next results, the identification algorithm is applied to an experimental data set of input-output values acquired from the implemented system described in preceding simulations. The system is controlled by an optimal law described in [9] and a load is connected to it. During this test, the motor speed has to follow a step reference. The acquired data set is composed by values of $iq_{ref}$ and the corresponding speed values.

Figure 10 displays the motor speed signal in response to a step reference. Until about 3s, a load is applied to the motor. After, the load is desacoupled and the controller has to generate a current reference in the manner to guide the motor speed to the reference signal.

The identification algorithm is applied to extract the relational matrix composed by the relationships between current reference $iq_{ref}$ and the motor speed. The algorithm uses the following parameters: $\alpha$ = 2.3 for the gravity-center's adjustment, $\beta$ = 0.85 in the adjustment of each universe of discourse, and a filter of order 1 with $\gamma$ =0.01. The results of one-step forward prediction are shown in figure 11 revealing the good performance obtained with the proposed relational algorithm in predict the motor speed. Figure 12 shows the error signal between real and predicted speed values. Notice that during the initial instants the algorithm is builting the relational matrix and the error has high values. After, as the model relations are extracted, the prediction error decreases approximating the predicted speed signal with the measured one.

## VII. CONCLUSION

We have proposed in this paper a new fuzzy relational identification algorithm and presented its use in predicting the speed signal of a motor drive system.

The algorithm is an extension of an initially proposed relational identification algorithm [1] in which a noise filtering technique and a new version of the gravity-center's adjustment was described.

The fuzzy algorithm was applied to a motor drive system with simulated and experimental results for different situations. The results demonstrated its potentialities in filter the noise influence into the extracted relational matrix and the good prediction of motor drive’s speed.

# APPENDIX

## MOTOR NOMENCLATURE

| | |
|---|---|
| $i_d , i_q$ | Stator currents described in *d*-*q* components. |
| $L_d , L_q$ | Self-induction coefficients in *d*-*q* components. |
| $u_d , u_q$ | Stator voltages in *d*-*q* components. |
| $R$ | Stator resistance. |
| $\phi_f$ | Magnetic flux associated with the rotor. |
| $T_x$ | Load. |
| $\theta$ | Rotor position. |
| $\omega$ | Angular speed. |

## MACHINE PARAMETERS

| |
|---|
| $R = 0.9\ \Omega$<br>$\phi_f = 0.5\ \text{Wb}$ |
| $L_d = 0.0051\ \text{H}$<br>$L_q = 0.0056\ \text{H}$ |
| $U = 250\ \text{V}$<br>$J = 0.025\ \text{Kg.m}^2$ |

Fig. 1. Three types of mostly used fuzzy sets.

Fig. 2. Membership functions.

Fig. 3. Block diagram of the identification algorithm.

Fig. 4. Permanent-Magnet Drive System used in this study.

Fig. 5. Reference current $iq_{ref}$ (input-signal).

Fig. 6. Real (doted line) and predicted (solid line) motor speed signals.

Fig. 7. Error signal evolution ($\alpha$ = 2.3, $\beta$ = 0.82, $\gamma$ =0.01).

Fig. 8. Clean (dotted line) and filtered (solid line) speed signals for a filter of order $n = 1$ and parameters $\alpha$ = 1.4, $\beta$ = 0.9, and $\gamma$ =0.15.

Fig. 9. Clean (dotted line) and filtered (solid line) signals for a filter of order $n = 2$ and parameters $\alpha$ = 0.35, $\beta$ = 0.9, and $\gamma$ =0.35.

Fig. 10. Experimental motor speed signal obtained for a step reference.

Fig. 11. Predicted motor speed value ($\alpha$ = 2.3, $\beta$ = 0.85, $\gamma$ =0.01) as the identification algorithm extracts the relational matrix.

Fig. 13. Error between real and step-forward motor speed prediction.

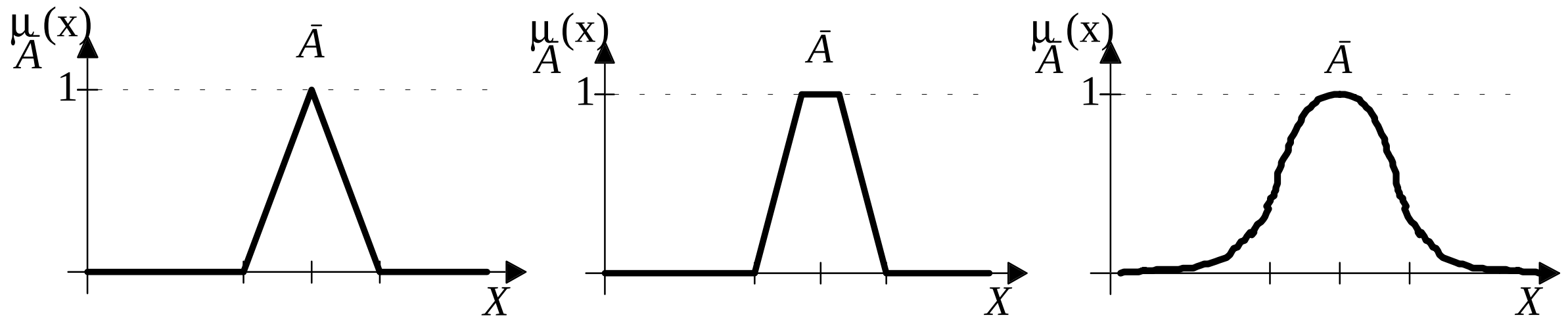

**Triangular** **Trapezoidal** **Gaussian**

**Figure 1**

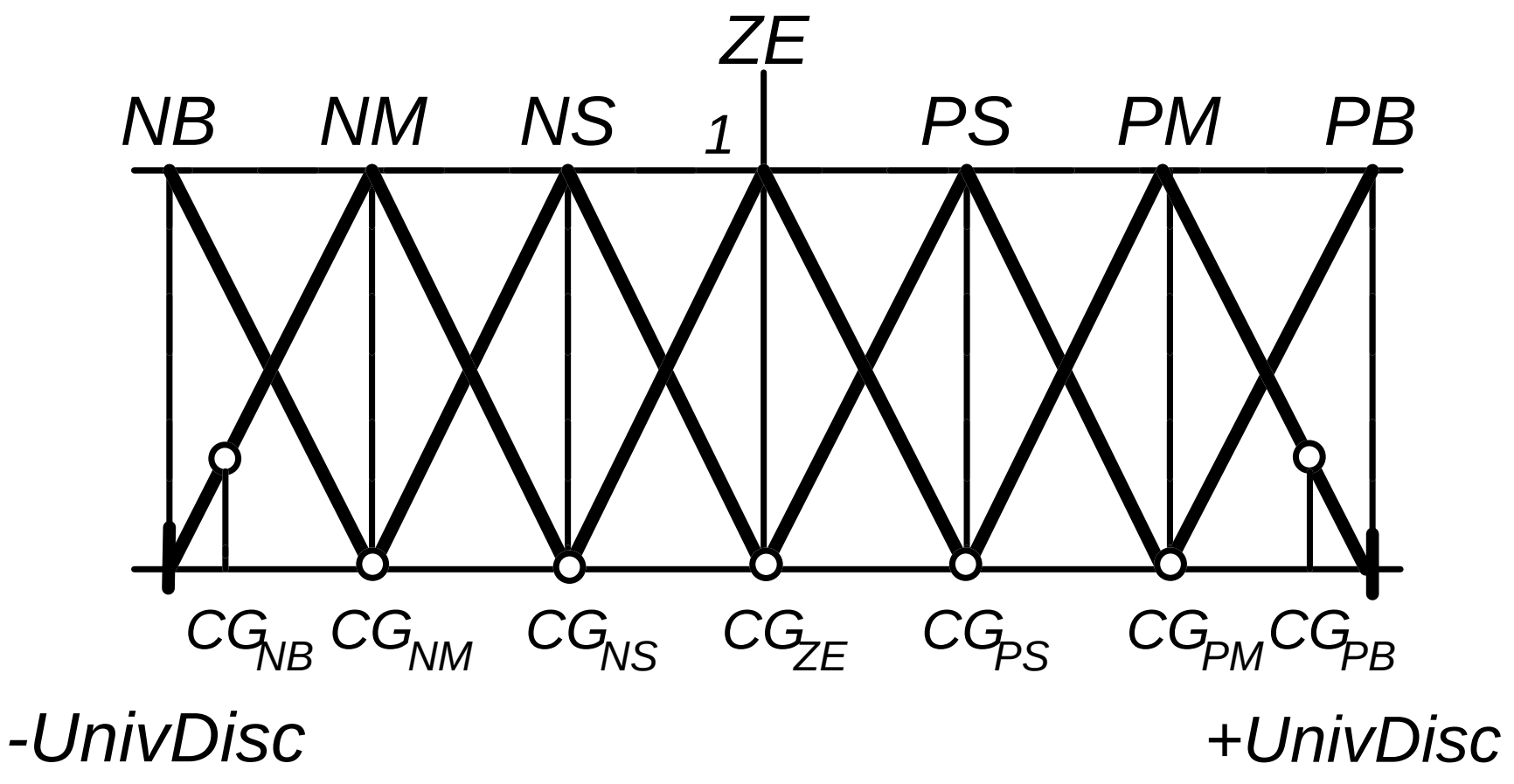

**Figure 2**

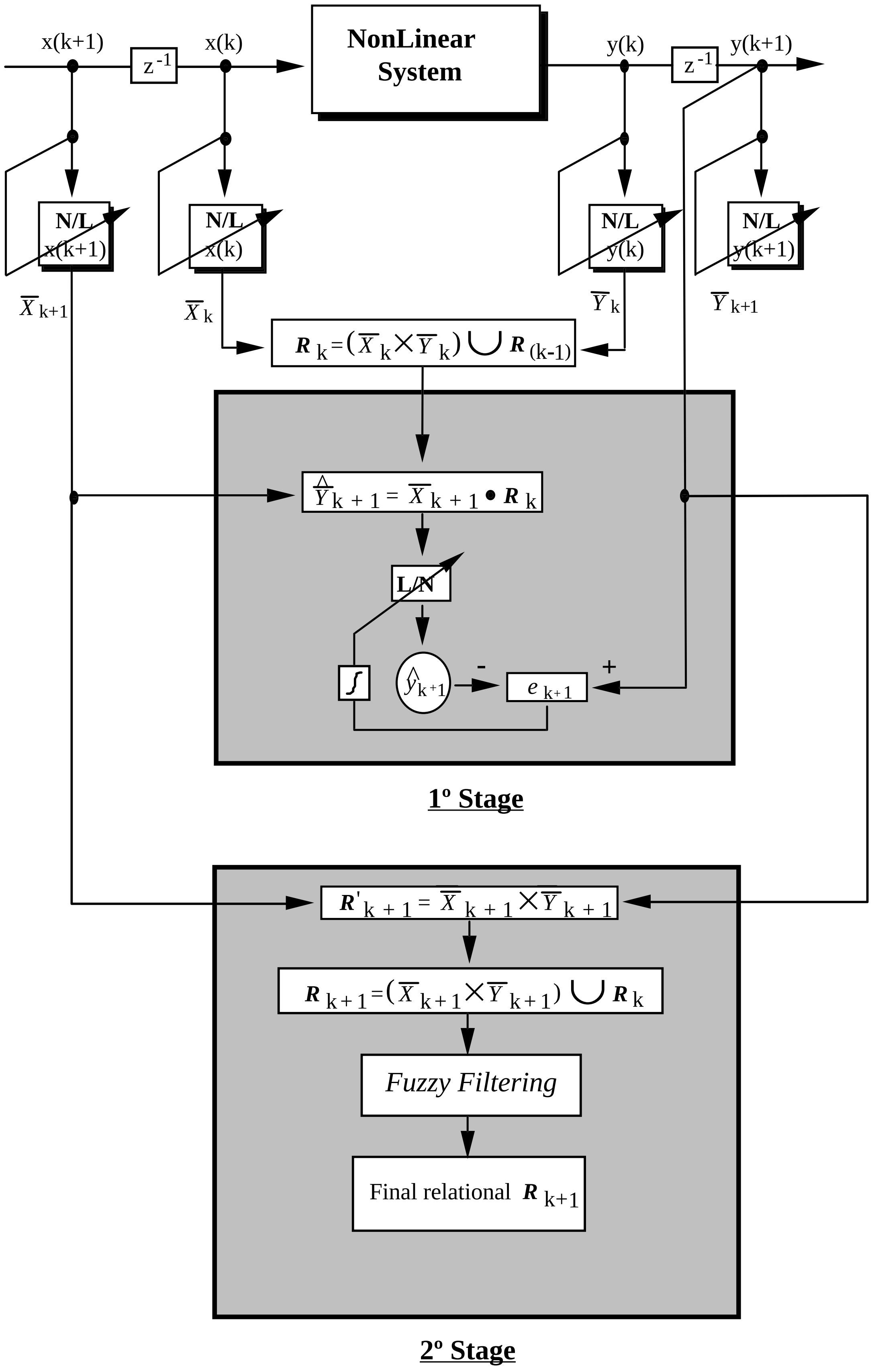
x(k+1)
z-1
x(k)
NonLinear System
y(k)
z-1
y(k+1)
N/L x(k+1)
N/L x(k)
N/L y(k)
N/L y(k+1)
X k+1
X k
Y k
Y k+1
R k = (X k × Y k) ∪ R (k-1)
Y k+1 = X k+1 • R k
L/N
y k+1
-
+
e k+1
1º Stage
R' k+1 = X k+1 × Y k+1
R k+1 = (X k+1 × Y k+1) ∪ R k
Fuzzy Filtering
Final relational R k+1
2º Stage

**Figure 3**

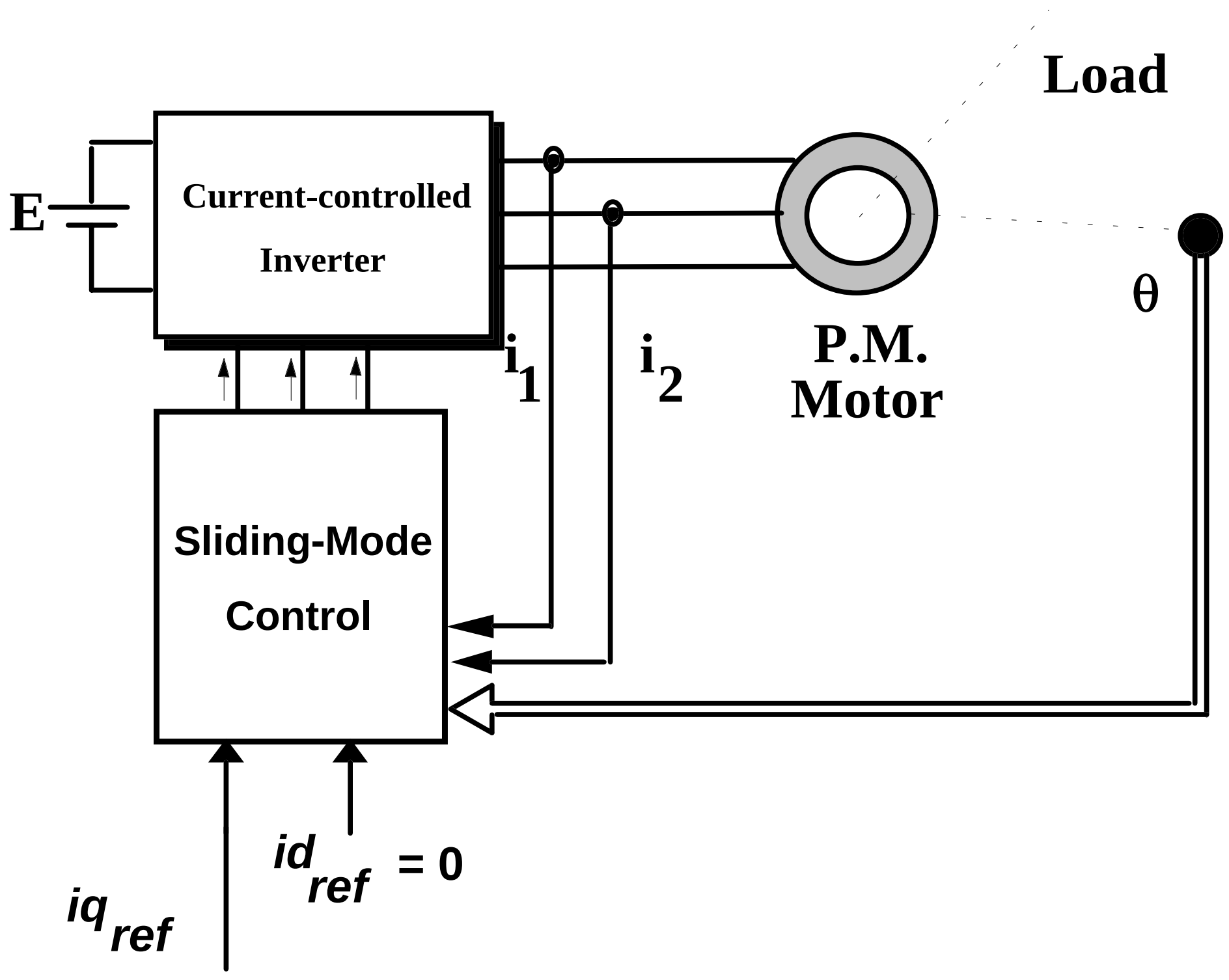

**Figure 4**

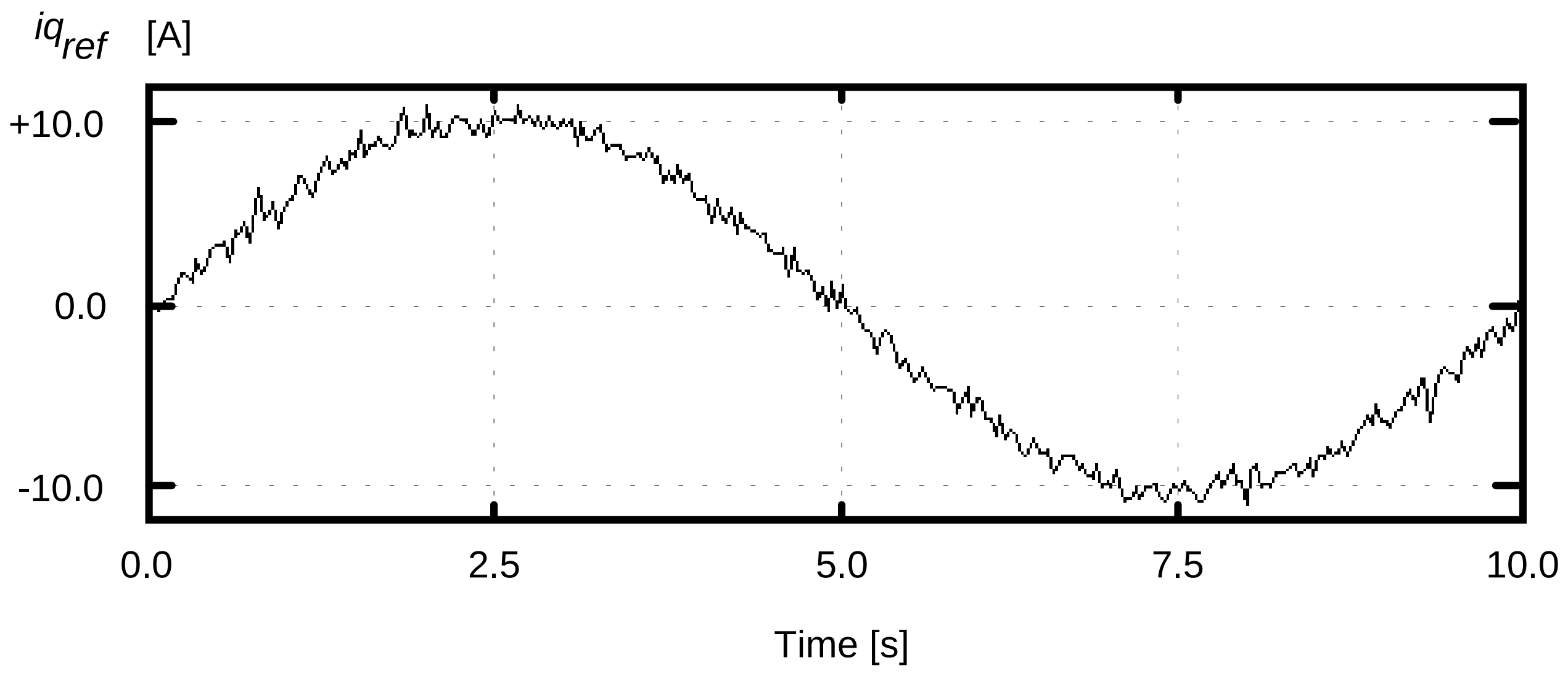

**Figure 5**

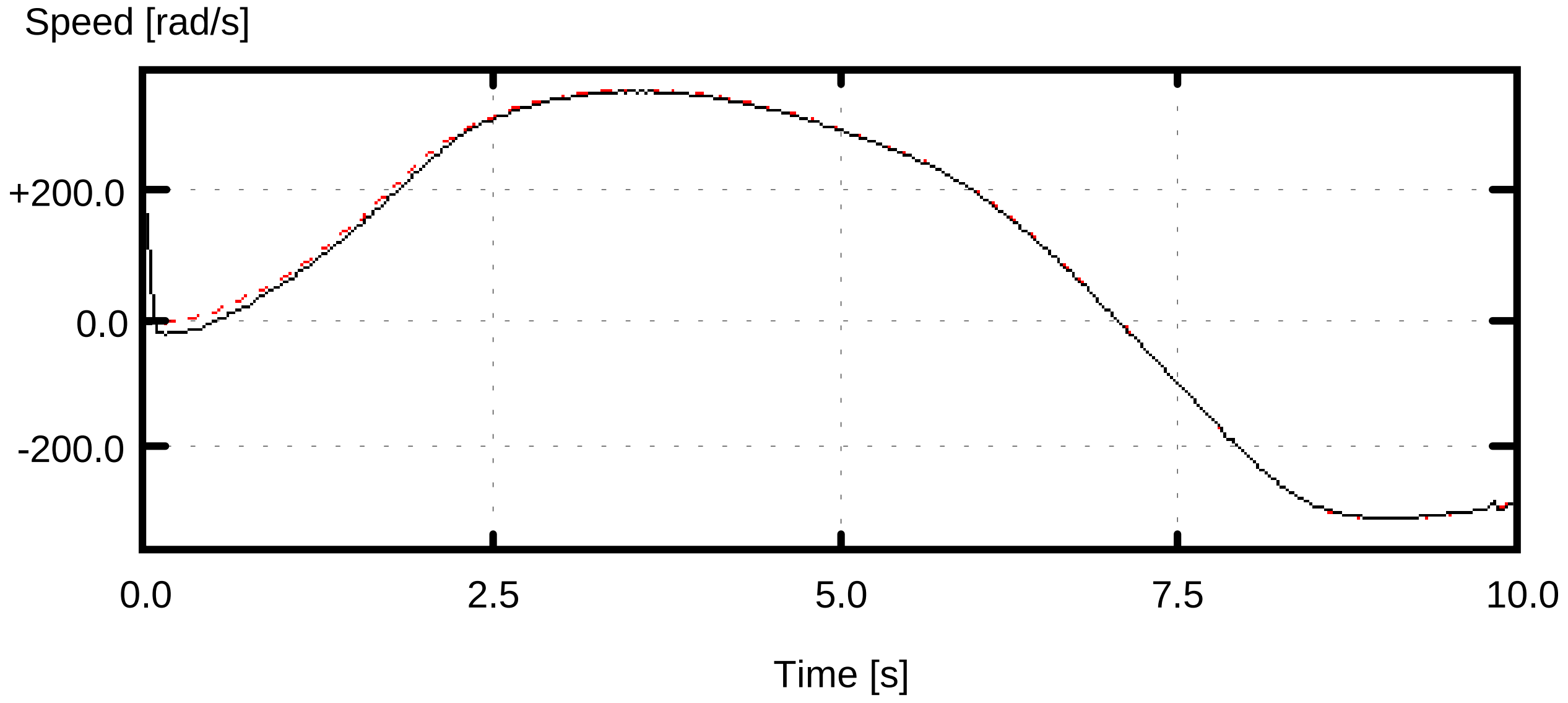

**Figure 6**

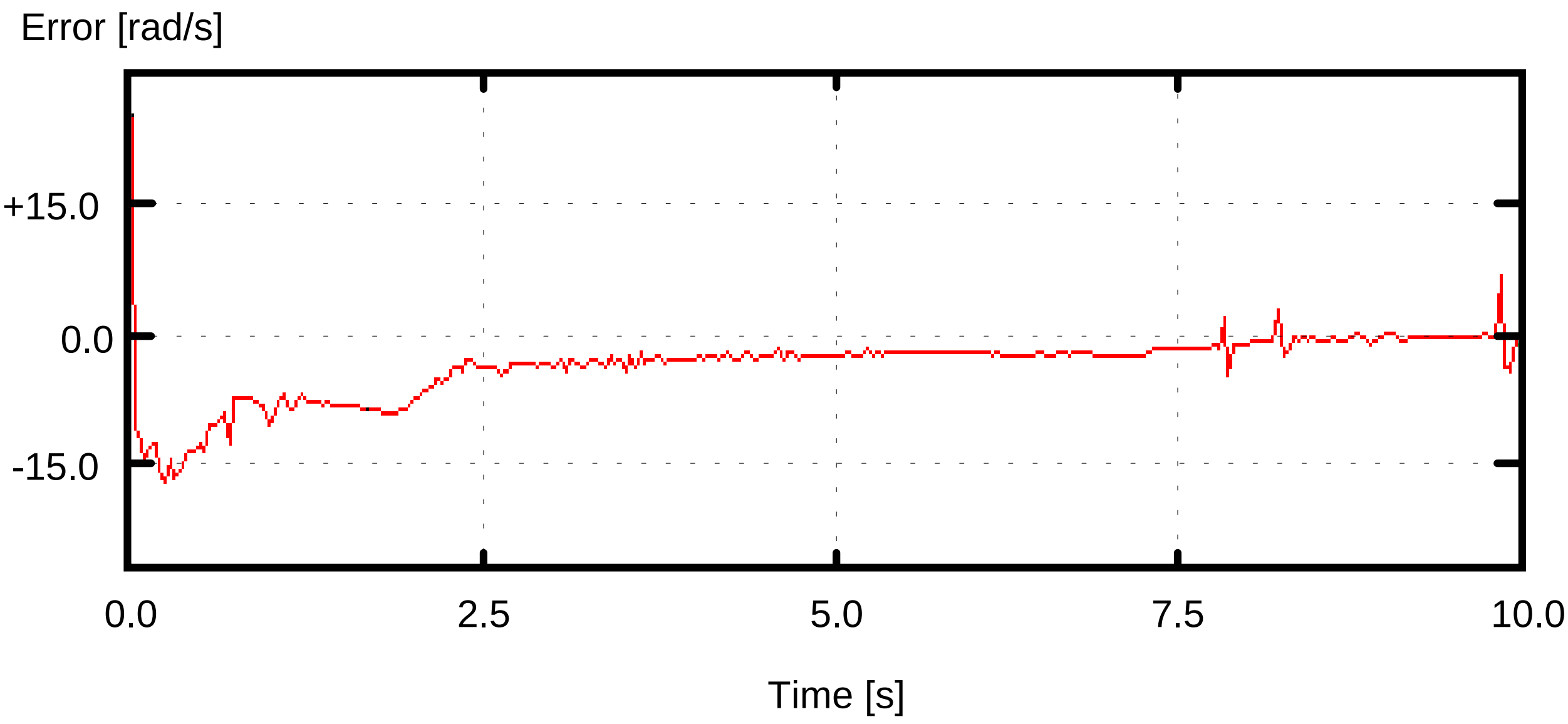

**Figure 7**

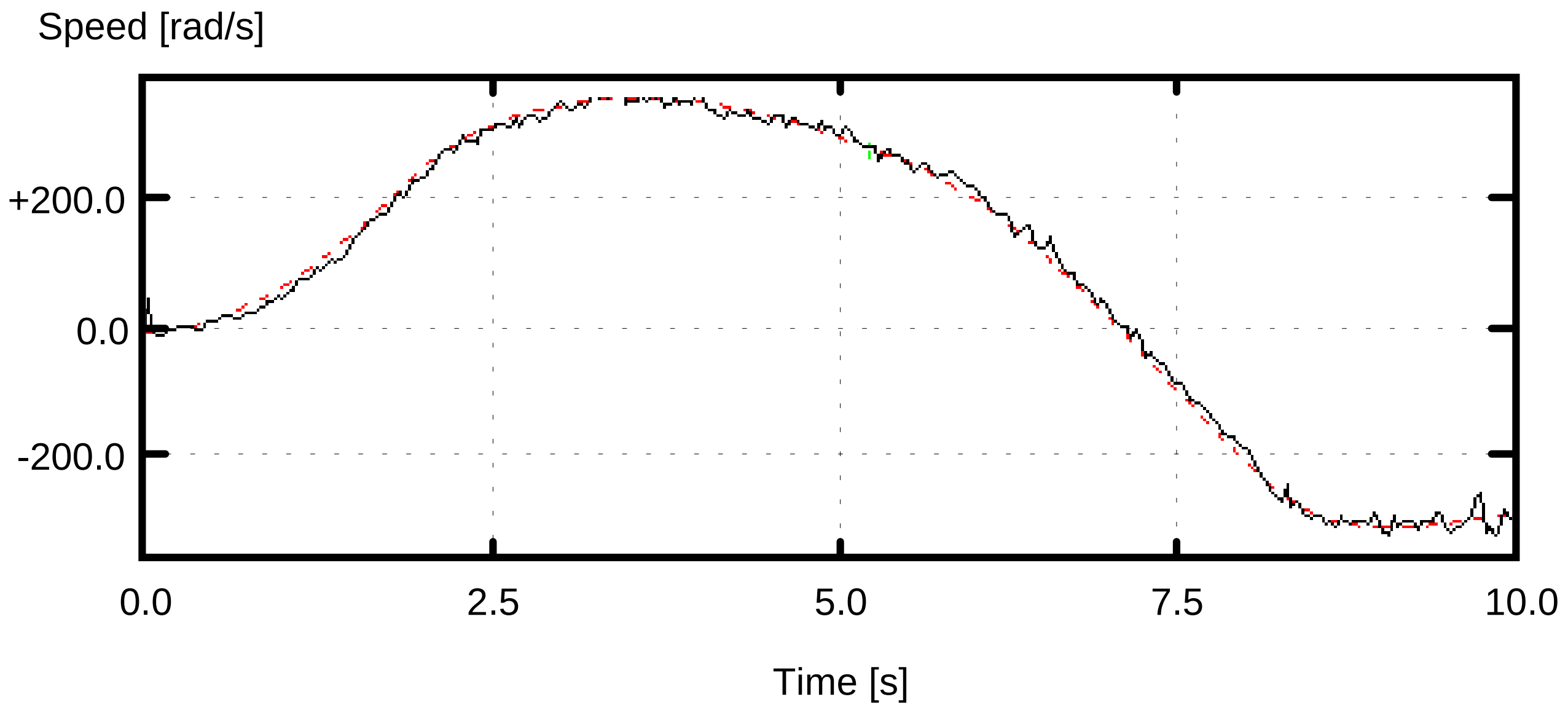

**Figure 8**

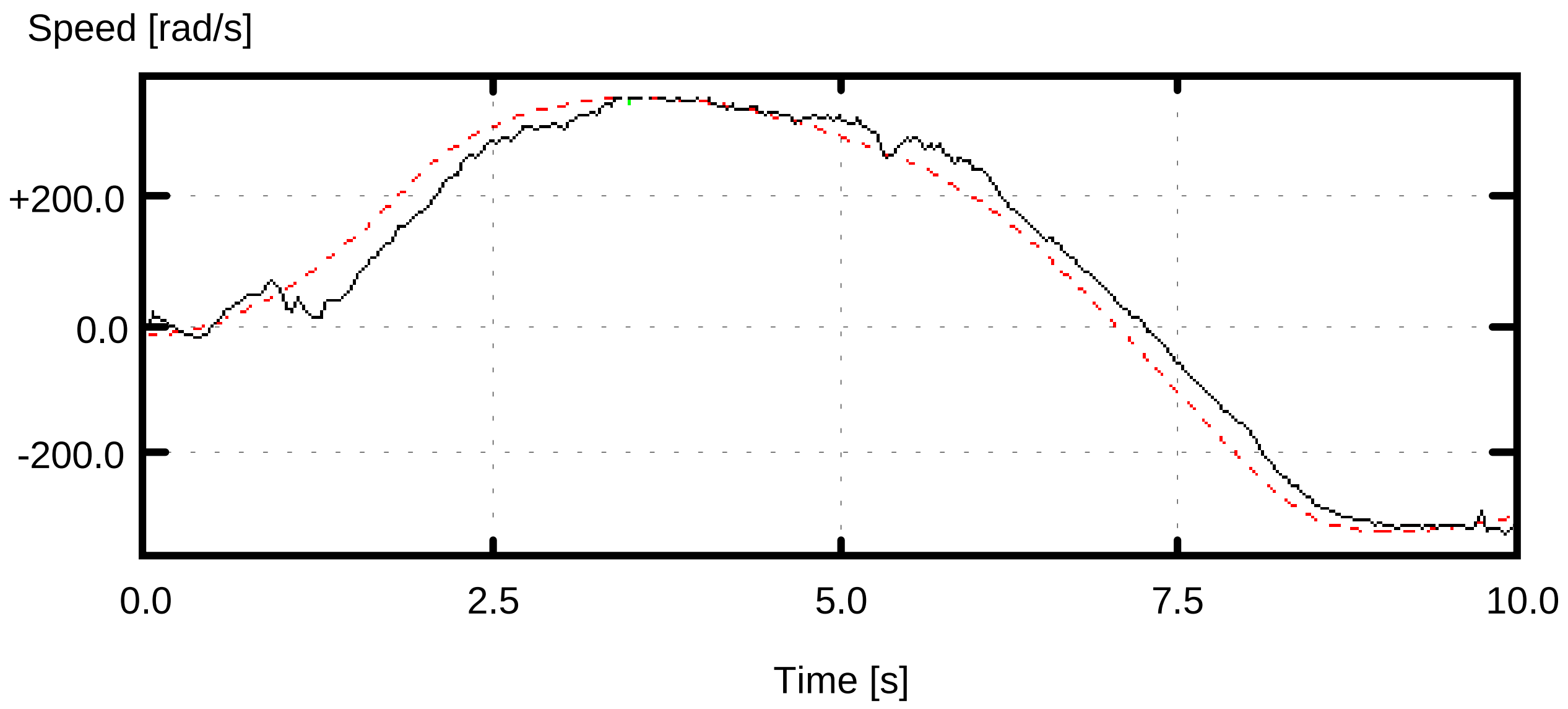

**Figure 9**

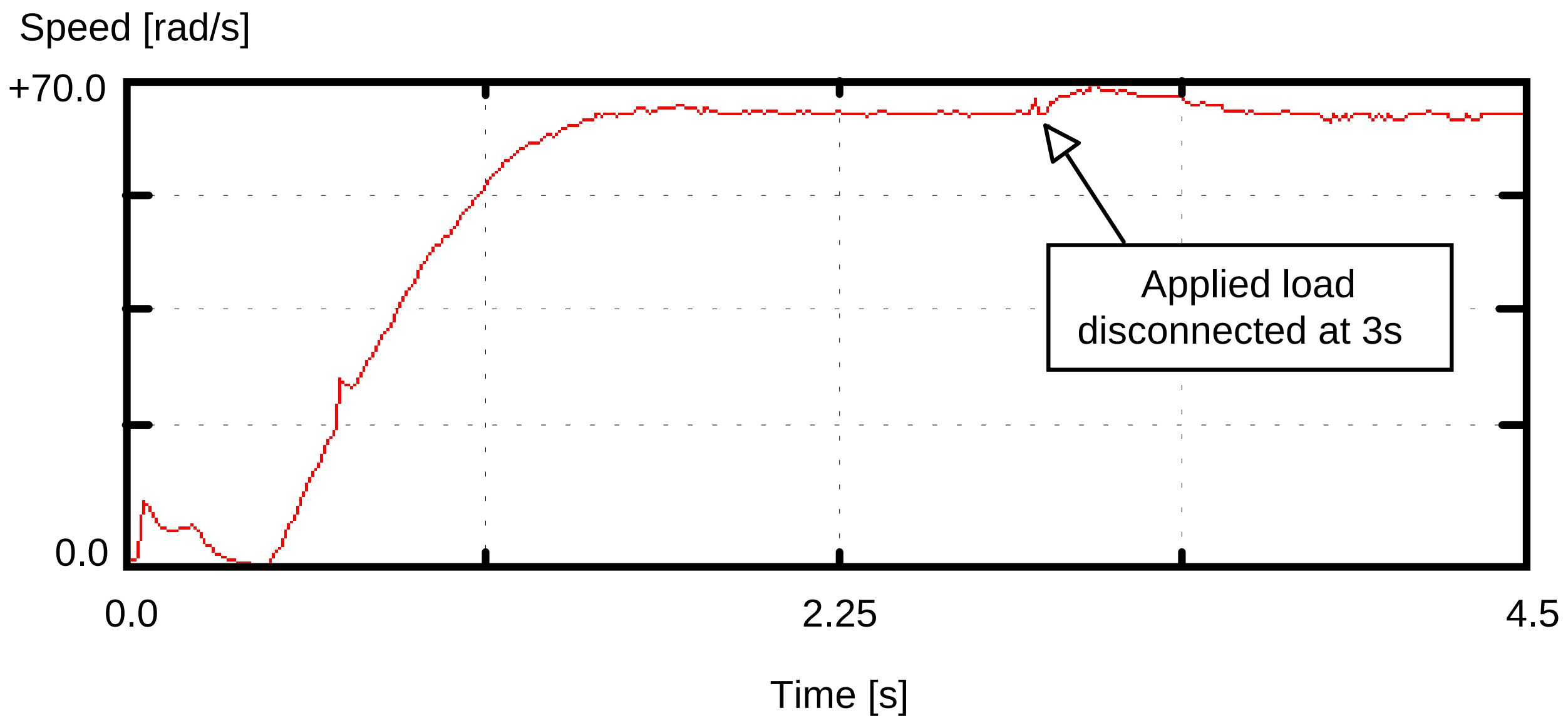

**Figure 10**

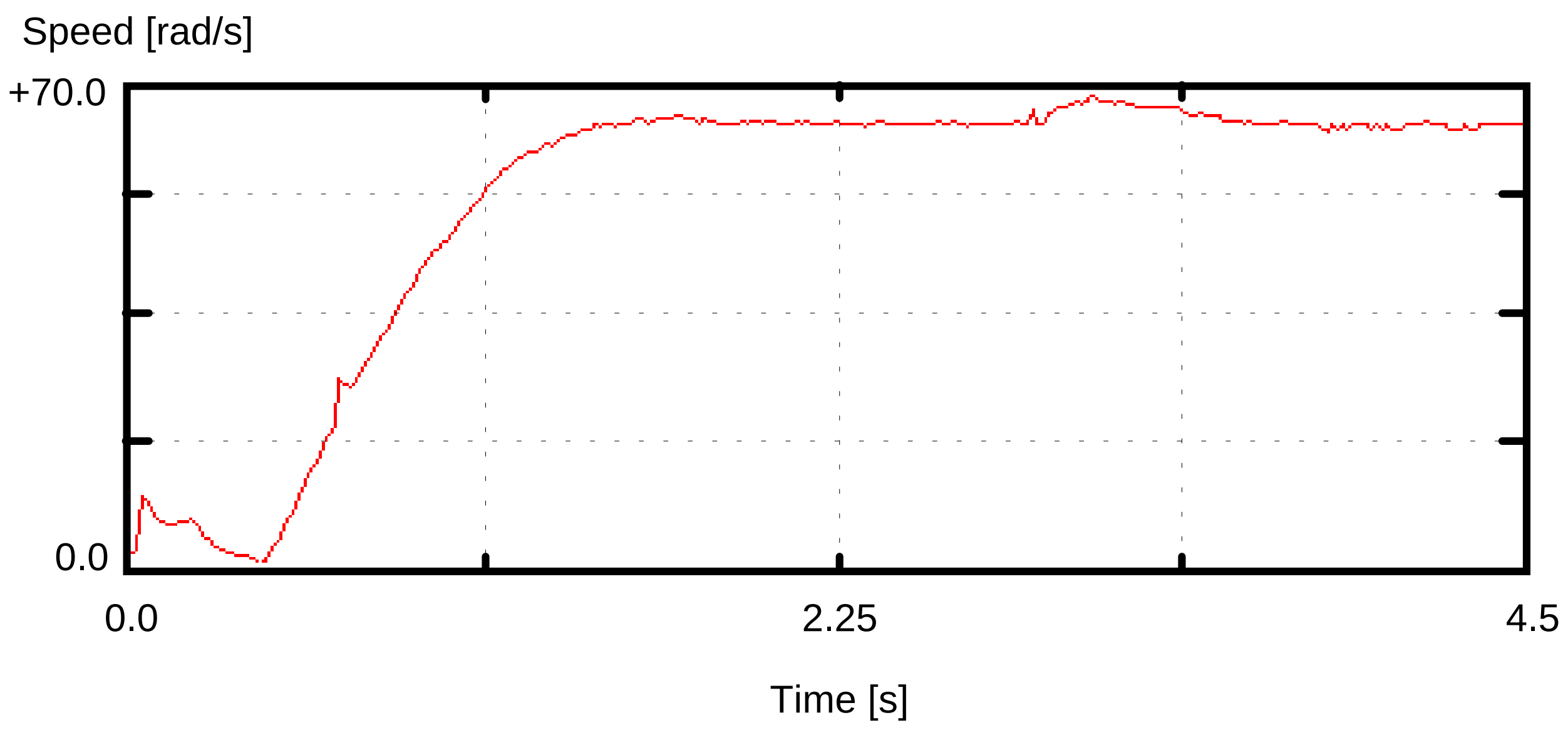

**Figure 11**

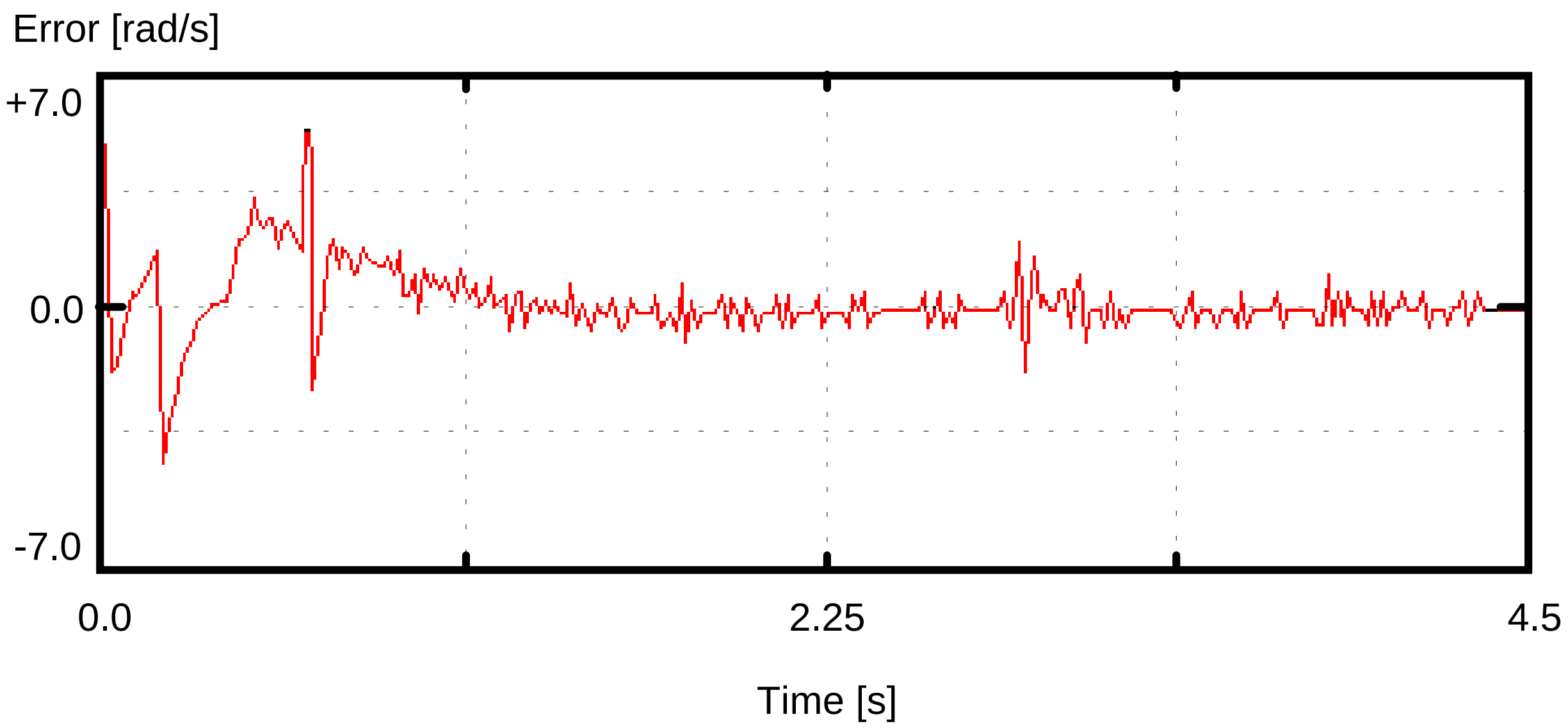

**Figure 12**